%% file: root.tex
\documentclass[letterpaper, 10 pt, conference]{ieeeconf}
\IEEEoverridecommandlockouts
\usepackage{cite}
\usepackage{amsmath,amssymb,amsfonts}
\usepackage{algorithmic}
\usepackage{graphicx}
\usepackage{textcomp}
\usepackage{xcolor}
\usepackage{multirow}
\def\BibTeX{{\rm B\kern-.05em{\sc i\kern-.025em b}\kern-.08em
    T\kern-.1667em\lower.7ex\hbox{E}\kern-.125emX}}

\title{\LARGE \bf
Variable Weights Neural Network For Diabetes Classification
}

\author{Tanmay Rathi$^{1}$ and Vipul$^{2}$
\thanks{*This work was not supported by any organization and both authors contributed equally}
\thanks{$^{1}$Tanmay Rathi with the Department of Electrical Engineering,
        Indian Institute of Technology Roorkee, Roorkee, INDIA
        {\tt\small trathi@ee.iitr.ac.in}}%
\thanks{$^{2}$Vipul with the Department of Electrical Engineering, Indian Institute of Technology Roorkee, Roorkee, INDIA 
        {\tt\small vipul@ee.iitr.ac.in}}%
}

\begin{document}
\maketitle

\begin{abstract}
As witnessed in the past year, where the world was brought to the ground by a pandemic, fighting Life-threatening diseases have found greater focus than ever. The first step in fighting a disease is to diagnose it at the right time. Diabetes has been affecting people for a long time, and is growing among people faster than ever. The number of people who have Diabetes reached 422 million in 2018, as reported by WHO, and the global prevalence of Diabetes among adults above the age of 18 has risen to 8.5\%. 
Now Diabetes is a disease that shows no or very few symptoms among the people affected by it for a long time, and even in some cases, people realize they have it when they have lost any chance of controlling it. So getting Diabetes diagnosed at an early stage can make a huge difference in how one can approach curing it. Moving in this direction in this paper, we have designed a liquid machine learning\cite{b2} approach to detect Diabetes with no cost using deep learning. In this work, we have used a dataset of 520 instances from \cite{b1}. Our approach shows a significant improvement in the previous state-of-the-art results. Its power to generalize well on small dataset deals with the critical problem of lesser data in medical sciences. 
\end{abstract}

\input{introduction}
\input{architecture}
\input{experiments}
\input{conclusion}
\input{future_work}

\end{document}

%% file: introduction.tex
\section{Introduction}
Diabetes mellitus has been around and affecting the human race since even before medical science. For a long time, we have tried to conquer Diabetes by finding new and better treatments and understanding its behavior and patterns. It is mainly classified as Type 1 and Type 2 Diabetes, where Type 1 occurs due to the body's incapability to produce insulin or produce tiny amounts of it. This insufficiency of insulin occurs mainly in children and adolescents. This type had no cure for a very long time and would cause fatality within weeks or months, but all that changed after the invention of artificial insulin. Type 2 is the one most prevalent among adults and occurs due to inefficient use of the insulin produced inside the body. This type of Diabetes can originate and grow in a person's body without showing noticeable symptoms for over seven years. There is another type of Diabetes, Gestational Diabetes, which mainly develops in pregnant women due to hormonal changes. We have not yet been able to discover a perfect single cure for Diabetes, so the best we can do is reduce the risk of it as early as possible. Due to lack of extreme suggesting symptoms, it's medically not possible to confirm the disease. Still, it can be predicted to happen in the future for a person using the trends seen among those who suffer from it, even from very nominal symptoms.
The text will introduce a machine learning model to detect Diabetes at no cost from few simple queries. We used the dataset from \cite{b1} containing 520 instances of diabetic and non-diabetic people to train our model. We provided state-of-the-art results on this dataset. Despite its small size, we generalize well over the unseen data.

%% file: architecture.tex
\section{Architecture}

\begin{figure}[htbp]
\centerline{\includegraphics[width=90mm,scale=1]{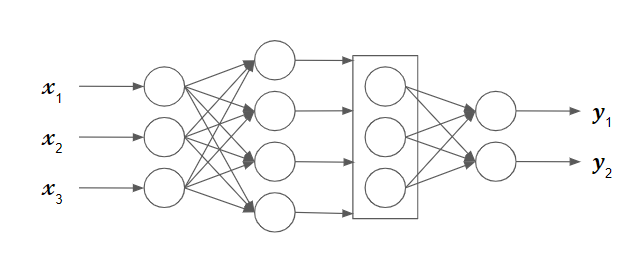}}
\caption{Variable Weight Layer can be used with standard neural network layer providing the additional benefit of Variability and Generalization.}
\label{fig1}
\end{figure}

We proposed a liquid machine learning\cite{b2} approach for better generalization, even with a small dataset. We call this layer simply a Variable Weight Layer. Our layer's remarkable features are 1) better generalization, 2) easy stacking with other layers, 3)  and full model development. The Variable Weight layer can provide decent generalizability even on small datasets. It can be stacked together to build a full architecture like a standard neural network layer or can be used with a standard neural network and provides an extra variability in weights for better generalization as in ``Fig.~\ref{fig1}''. The deriving idea behind the layer development is that neural networks have stationary weights that become constant ones trained, making it difficult for them to behave correctly in changing conditions. Our proposed layer decides its weights based on the given input and processes input accordingly, hence deciding how to process a particular input. \\

\begin{figure}[htbp]
\centerline{\includegraphics[width=90mm,scale=1]{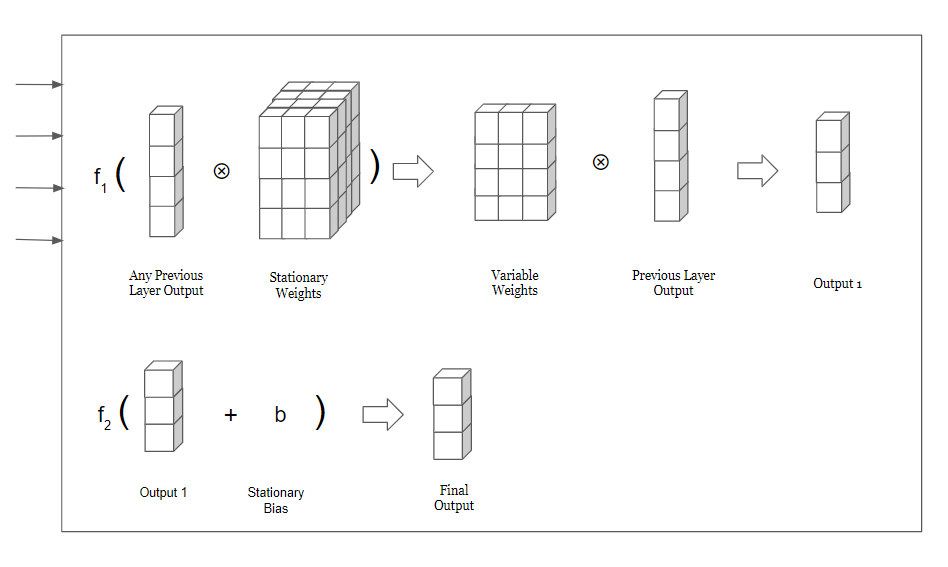}}
\caption{Weight Prediction and Use in Variable Weight Layer.}
\label{fig2}
\end{figure}

The proposed Variable Weight Layer is shown in ``Fig.~\ref{fig2}''. The inputs from any previous network layers first pass through the stationary wights to produce the variable weights. These varying weights are then used to create the final output with the static bias value as shown in ``Fig.~\ref{fig2}''. "f\textsubscript{1}" and "f\textsubscript{2}" are the two activation functions used. For "f\textsubscript{1}," since predicted weights should be bi-polar, we experimented with linear and Tanh activation functions. Finally, we concluded that Tanh works better and also provides weight normalization as an additional benefit. For "f\textsubscript{2}," we used the ReLU activation function. For more detailed comparisons between the deriving equation of a neural network layer, variable weight layer, and variable bias layer, refer to the results section.\\

One worth noting point here is the exponential increase in weights, which we can deal with by predicting only bias or a particular weight. Indeed, predicting only a single weight decreases the accuracy compared to the original layer but reduces the weight drastically and still better than the standard artificial neural network. As per our final version, we have also shown that even with a lot fewer parameters, we can generalize well than the standard artificial neural networks, proving that our changing weights can generalize well over the unseen data.

%% file: experiments.tex
\section{Experiments}
\subsection{Dataset}
The Diabetes dataset\cite{b1} contains 520 instances of diabetic and non-diabetic peoples with 16 attributes. The dataset is a classification dataset where the result is to classify a person as diabetic or not based on features given. We decided to go with this dataset as this is a medical dataset and contains very few training examples to train a big network. Which correctly represents the current scenario in medical sciences. After preprocessing, our dataset finally has 500 instances, out of which 314 are non-diabetic people, and 186 are diabetic. Tables I and II summarise the dataset statistics.The data has been collected from the patients using a direct questionnaire from Sylhet Diabetes Hospital of Sylhet, Bangladesh. 

\begin{table}[htbp]
\caption{Description of Dataset}
\resizebox{\columnwidth}{!}{%
\begin{tabular}{|c|c|c|}
\hline
\textbf{}                                                           & Number of Attributes & Number of Instances \\ \hline
\begin{tabular}[c]{@{}c@{}}Diabetes Symptom \\ Dataset\end{tabular} & 16                   & 520                 \\ \hline
\end{tabular}}
\end{table}

\begin{table}[htbp]
\caption{Description of Attributes}
\resizebox{\columnwidth}{!}{%
\begin{tabular}{cc}
\hline
\multicolumn{1}{|c|}{\textbf{Attributes}} & \multicolumn{1}{c|}{\textbf{Values}}                                \\ \hline
\multicolumn{1}{|c|}{Age}                 & \multicolumn{1}{c|}{1.20-35, 2.36-45, 3.46-55,4.56-65, 6 above 65.} \\ \hline
\multicolumn{1}{|c|}{Sex}                 & \multicolumn{1}{c|}{1.Male, 2.Female.}                              \\ \hline
\multicolumn{1}{|c|}{Polyuria}            & \multicolumn{1}{c|}{1.Yes, 2.No.}                                   \\ \hline
\multicolumn{1}{|c|}{Polydipsia}          & \multicolumn{1}{c|}{1.Yes, 2.No.}                                   \\ \hline
\multicolumn{1}{|c|}{sudden weight loss}  & \multicolumn{1}{c|}{1.Yes, 2.No.}                                   \\ \hline
\multicolumn{1}{|c|}{Weakness}            & \multicolumn{1}{c|}{1.Yes, 2.No.}                                   \\ \hline
\multicolumn{1}{|c|}{Polyphagia}          & \multicolumn{1}{c|}{1.Yes, 2.No.}                                   \\ \hline
\multicolumn{1}{|c|}{Genital thrush}      & \multicolumn{1}{c|}{1.Yes, 2.No.}                                   \\ \hline
\multicolumn{1}{|c|}{visual blurring}     & \multicolumn{1}{c|}{1.Yes, 2.No.}                                   \\ \hline
\multicolumn{1}{|c|}{Itching}             & \multicolumn{1}{c|}{1.Yes, 2.No.}                                   \\ \hline
\multicolumn{1}{|c|}{Irritability}        & \multicolumn{1}{c|}{1.Yes, 2.No.}                                   \\ \hline
\multicolumn{1}{|c|}{delayed healing}     & \multicolumn{1}{c|}{1.Yes, 2.No.}                                   \\ \hline
\multicolumn{1}{|c|}{partial paresis}     & \multicolumn{1}{c|}{1.Yes, 2.No.}                                   \\ \hline
\multicolumn{1}{|c|}{muscle stiffness}    & \multicolumn{1}{c|}{1.Yes, 2.No.}                                   \\ \hline
\multicolumn{1}{|c|}{Alopecia}            & \multicolumn{1}{c|}{1.Yes, 2.No.}                                   \\ \hline
\multicolumn{1}{|c|}{Obesity}             & \multicolumn{1}{c|}{1.Yes, 2.No.}                                   \\ \hline
\multicolumn{1}{|c|}{Class}             & \multicolumn{1}{c|}{1.Positive, 2.Negative.}                                   \\ \hline
\end{tabular}}
\end{table}
\subsection{Results}
 The performance of different algorithms on the Diabetes dataset is given in Tables III and IV. The previous state-of-the-art method on this dataset was Random Forest Algorithm. Using ten fold cross-validation, 97.4\% of instances were classified correctly, and using the percentage split technique could classify 99\% of the instances correctly. We first tested a standard neural network and achieved a similar 97.4\% cross-validation score and a 99\% percentage split. With our newly designed layer, we achieved a 99.4\% cross-validation score and 100\% percentage-split score. Compared with the standard neural network, our freshly designed layer generalizes well and increases 2\% on cross-validation and 1\% percent on the percentage split. To reduce the weight overhead, we only used our layer at the end of the network. To reduce the parameters further, we train a network with half of the parameters and achieve a 1\% increase in cross-validation and percentage split metric. To achieve such a decrease in parameters, we used our newly designed layer configured to have variable biases rather than variable weights.\\\\
The equations behind the three layers we compared above are - \\
\input{equ}

Acronyms used in Tables III and IV are NB - Naive Bayes, LR - Logistic Regression, J48 Decesion Tree, RF - Random forest, NN - Neural Network, VB - Variable Bias Network, VW - Varible Weight Network 

\begin{table}[hbtp]
\caption{Comparison Of Evaluation Metrices Using 10 Fold Cross-Validation}
\resizebox{\columnwidth}{!}{%
\begin{tabular}{|c|c|c|c|c|c|c|c|}
\hline
\multirow{2}{*}{\begin{tabular}[c]{@{}c@{}}Evaluation \\ Metrics\end{tabular}}       & \multicolumn{7}{c|}{Cross-Validation}                        \\ \cline{2-8} 
                                                                                     & NB     & LR     & J48    & RF     & NN     & VB     & VW     \\ \hline
\begin{tabular}[c]{@{}c@{}}Total Number \\ of \\ Instances\end{tabular}              & 500    & 500    & 500    & 500    & 500    & 500    & 500    \\ \hline
\multirow{2}{*}{\begin{tabular}[c]{@{}c@{}}Correct\\ Classifications\end{tabular}}   & 437    & 462    & 478    & 487    & 487    & 492    & 497    \\ \cline{2-8} 
                                                                                     & 87.4\% & 92.4\% & 95.6\% & 97.4\% & 97.4\% & 98.4\% & 99.4\% \\ \hline
\multirow{2}{*}{\begin{tabular}[c]{@{}c@{}}Incorrect\\ Classifications\end{tabular}} & 63     & 38     & 22     & 13     & 13     & 8      & 3      \\ \cline{2-8} 
                                                                                     & 12.6\% & 7.6\%  & 4.4\%  & 2.6\%  & 2.6\%  & 1.6\%  & 0.6\%  \\ \hline
\end{tabular}}
\end{table}

\begin{table}[hbtp]
\caption{Comparison of Evaluation Metrices Using Percentage Split(80:20)}
\resizebox{\columnwidth}{!}{%
\begin{tabular}{|c|c|c|c|c|c|c|c|}
\hline
\multirow{2}{*}{Evaluation Metrics}                                                  & \multicolumn{7}{c|}{Percentage Split}            \\ \cline{2-8} 
                                                                                     & NB   & LR   & J48  & RF   & NN   & VB    & VW    \\ \hline
\begin{tabular}[c]{@{}c@{}}Total Number \\ of \\ Instances\end{tabular}              & 100  & 100  & 100  & 100  & 100  & 100   & 100   \\ \hline
\multirow{2}{*}{\begin{tabular}[c]{@{}c@{}}Correct\\ Classifications\end{tabular}}   & 88   & 91   & 95   & 99   & 99   & 100   & 100   \\ \cline{2-8} 
                                                                                     & 88\% & 91\% & 95\% & 99\% & 99\% & 100\% & 100\% \\ \hline
\multirow{2}{*}{\begin{tabular}[c]{@{}c@{}}Incorrect\\ Classifications\end{tabular}} & 12   & 9    & 5    & 1    & 1    & 0     & 0     \\ \cline{2-8} 
                                                                                     & 12\% & 9\%  & 5\%  & 1\%  & 1\%  & 0\%   & 0\%   \\ \hline
\end{tabular}}
\end{table}

%% file: equ.tex
1) Standard Neural Network layer -\\\\
In standard Neural Networks, we use the previous layer outputs $l^{i-1}$ with the current layers stationary weights $W_{s}$ and biases $b_{s}$ to produce the output of the next layer $l^{i-1}$ as in below-given equations.    
Here $l_{j}^i$ is the $j^{th}$ output of $i^{th}$ layer. $\widehat{g}$ is the ReLU activation function.
\\\\
\begin{math}
l_{1}^{(i)} = \widehat{g}(\,l_{1}^{(i-1)}*w_{s_{11}} + l_{2}^{(i-1)}*w_{s_{21}} +\ldots+ l_{n}^{(i-1)}*w_{s_{n1}} + b_{\widehat{s}_{1}}\,)\\
\\
l_{2}^{(i)} = \widehat{g}(\,l_{1}^{(i-1)}*w_{s_{12}} + l_{2}^{(i-1)}*w_{s_{22}} +\ldots+ l_{n}^{(i-1)}*w_{s_{n2}} + b_{\widehat{s}_{2}}\,)\\
\\
\vdots
\\
\\
l_{n}^{(i)} = \widehat{g}(\,l_{1}^{(i-1)}*w_{s_{1n}} + l_{2}^{(i-1)}*w_{s_{2n}} +\ldots+ l_{n}^{(i-1)}*w_{s_{nn}} + b_{\widehat{s}_{n}}\,)\\
\\
l^{(i)} = \begin{bmatrix}
l_{1}^{(i)} &l_{2}^{(i)}  &\ldots  &l_{n}^{(i)} 
\end{bmatrix}
\end{math}
\\
\\\\
2) Variable Weight Layer - \\\\
In variable weight layer we take the input from any previous layer of the network $p = l^k,k\,\epsilon\,[1,i-1]$ where $l^i$ is the current layer and use it to predict the variable weights $W_{v}$ using stationary weights $W_{s}$ and biases $b_{s}$. The predicted variable weights assist further in the final output. $f_{1}$ is Tanh activation function and $\widehat{f_{2}}$ is ReLU. In our results $p = l^{(i-1)}$.
\\\\ 
\begin{math}
w_{v_{11}} = f_{1}(\,p_{1}*W_{s_{111}} + p_{2}*W_{s_{112}} + \ldots + p_{n}*W_{s_{11n}} + b_{s_{11}}\,)\\
\\
w_{v_{12}} = f_{1}(\,p_{1}*W_{s_{121}} + p_{2}*W_{s_{122}} + \ldots + p_{n}*W_{s_{12n}} + b_{s_{12}}\,)\\
\\
\vdots
\\
w_{v_{nn}} = f_{1}(\,p_{1}*W_{s_{nn1}} + p_{2}*W_{s_{nn2}} + \ldots + p_{n}*W_{s_{nnn}} + b_{s_{nn}}\,)\\
\\
W_{v} = \begin{bmatrix}
 w_{v_{11}}& w_{v_{12}}  & \ldots  &w_{v_{1n}} \\ 
 w_{v_{21}}& w_{v_{22}} & \ldots & w_{v_{2n}}\\ 
 \vdots & \vdots & \ddots & \vdots\\ 
 w_{v_{n1}}& w_{v_{n2}}  & \ldots & w_{v_{nn}}
\end{bmatrix}
\\\\
\\l_{1}^{(i)} = \widetilde{f_{2}}(\,p_{1}*w_{v_{11}} + p_{2}*w_{v_{21}} +\ldots+ p_{n}*w_{v_{n1}} + b_{\widetilde{s}_{1}}\,)\\
\\
l_{2}^{(i)} = \widetilde{f_{2}}(\,p_{1}*w_{v_{12}} + l_{2}*w_{v_{22}} +\ldots+ l_{n}*w_{v_{n2}} + b_{\widetilde{s}_{2}}\,)\\
\\
\vdots
\\
\\
l_{n}^{(i)} = \widetilde{f_{2}}(\,p_{1}*w_{v_{1n}} + p_{2}*w_{v_{2n}} +\ldots+ p_{n}*w_{v_{nn}} + b_{\widetilde{s}_{n}}\,)\\
\\
l^{(i)} = \begin{bmatrix}
l_{1}^{(i)} &l_{2}^{(i)}  &\ldots  &l_{n}^{(i)} 
\end{bmatrix}
\\
\end{math}
\\
3) Variable Bias Layer - \\\\
Variable Bias Layer contains a lot lesser parameters as we are only predicting the bias values. We can also predict $w_{s_{1\,1-n}}$ or any other weights and make them variable, giving us the freedom to decide the weight overhead. $g$ is linear activation here, and $\widehat{g}$ is ReLU. 
\\\\
\begin{math} 
b_{v_{1}} = g(\,p_{1}*W_{s_{11}} + p_{2}*W_{s_{21}} + \ldots + p_{n}*W_{s_{n1}} + b_{s_{1}}\,)\\
\\
b_{v_{2}} = g(\,p_{1}*W_{s_{12}} + p_{2}*W_{s_{22}} + \ldots + p_{n}*W_{s_{n2}} + b_{s_{2}}\,)\\
\\
b_{v_{n}} = g(\,p_{1}*W_{s_{1n}} + p_{2}*W_{s_{2n}} + \ldots + p_{n}*W_{s_{nn}} + b_{s_{n}}\,)\\
\\
b_{v} = 
\begin{bmatrix}
b_{v_{1}} &b_{v_{2}}  & \ldots &b_{v_{n}} 
\end{bmatrix}
\\
l_{1}^{(i)} = \widehat{g}(\,p_{1}*w_{s_{11}} + p_{2}*w_{s_{21}} +\ldots+ p_{n}*w_{s_{n1}} + b_{v_{1}}\,)\\
\\
l_{2}^{(i)} = \widehat{g}(\,p_{1}*w_{s_{12}} + p_{2}*w_{s_{22}} +\ldots+ p_{n}*w_{s_{n2}} + b_{v_{2}}\,)\\
\\
\vdots
\\
\\
l_{n}^{(i)} = \widehat{g}(\,p_{1}*w_{s_{1n}} + p_{2}*w_{s_{2n}} +\ldots+ p_{n}*w_{s_{nn}} + b_{v_{n}}\,)\\
\\
l^{(i)} = \begin{bmatrix}
l_{1}^{(i)} &l_{2}^{(i)}  &\ldots  &l_{n}^{(i)}
\end{bmatrix}
\\
\end{math}

%% file: conclusion.tex
\section{Conclusion}
With the rapid growth in people being affected by Diabetes, it has naturally become more important and necessary for it to be detected as early as possible to help cure it, and if figured out at a very early stage, then even some lifestyle changes can be sufficient, and one may not have to go through a lot of medications. Early symptoms such as sudden weight loss, obesity, muscle stiffness can help detect the possibility of having Diabetes through our modified MLP neural network model, which has achieved fantastic percentage split accuracy of 100\% and has proven to generalize well for new entries with a cross-validation score of 99.4\%. However, this model can be trained and modified with newly available data to perform even better, and some changes can be made to the model to have greater efficiency.

%% file: future_work.tex
\section{Future Directions}
In the model architecture number of parameters to be determined during model training increases exponentially with the number of features or neurons in the layer preceding the custom or dynamic layer. This will not be feasible for most classification data as they tend to have a large number of features, and this could turn out to be highly inefficient in some cases. So several ways can be tested to reduce this exponential factor to a more linear factor and still have the advantage of that dynamic layer, such as instead of predicting all the weights for a layer, it would be more efficient to predict only all the bias or some weights which could turn out to be more useful in the model as compared to others. A lot more experiments have to be conducted on many more diverse kinds of datasets, which would help generalize this approach to a lot more scenarios. We will soon develop a full report on applications and provide a library for public use.

%% file: root.bbl
\begin{thebibliography}{00}
\bibitem{b1} Islam, M M Faniqul \& Ferdousi, Rahatara \& Rahman, Sadikur \& Bushra, Humayra. (2020). Likelihood Prediction of Diabetes at Early Stage Using Data Mining Techniques. 10.1007/978-981-13-8798-2\_12.
\bibitem{b2} Ramin Hasani, Mathias Lechner, Alexander Amini, Daniela Rus, \& Radu Grosu. (2020). Liquid Time-constant Networks.
\end{thebibliography}
